\documentclass[10pt]{article}
\usepackage[letterpaper, margin=1in]{geometry} 
\usepackage{times} 
\usepackage{graphicx} 
\usepackage{caption} 
\usepackage[colorlinks=true,linkcolor=blue,urlcolor=blue,citecolor=blue]{hyperref} 
\usepackage{amsmath,amssymb,amsfonts} 
\usepackage{setspace} 
\usepackage[numbers,sort&compress]{natbib} 
\usepackage{tabularx, booktabs} 
\usepackage[framemethod=TikZ]{mdframed} 
\usepackage{changepage} 
\usepackage[export]{adjustbox} 
\usepackage{subcaption} 
\usepackage{dirtytalk} 

\title{\vspace{-2cm}\fontsize{14pt}{16.8pt}\selectfont\textbf{A Large Language Model Outperforms Other Computational Approaches to the High-Throughput Phenotyping of Physician Notes}}
\author{
    \normalsize Syed I. Munzir\textsuperscript{1}, Daniel B. Hier\textsuperscript{1,2}, Chelsea Oommen\textsuperscript{1}, Michael D. Carrithers\textsuperscript{1}\\
    \textsuperscript{1}\normalsize Department of Neurology and Rehabilitation, University of Illinois at Chicago, Chicago, USA\\
    \textsuperscript{2}\normalsize Kummer Institute, Missouri University of Science and Technology, Rolla MO, USA
}
\date{} 
\makeatletter
\let\ps@oldplain\ps@plain
\renewcommand\ps@plain{\ps@empty}
\makeatother

\begin{document}
\maketitle
\thispagestyle{empty} 
\pagestyle{empty}

\section*{Abstract}
\textit{High-throughput phenotyping, the automated mapping of patient signs and symptoms to standardized ontology concepts, is essential to gaining value from electronic health records (EHR) in the support of precision medicine. Despite technological advances, high-throughput phenotyping remains a challenge. This study compares three computational approaches to high-throughput phenotyping: a Large Language Model (LLM) incorporating generative AI, a Natural Language Processing (NLP) approach utilizing deep learning for span categorization, and a hybrid approach combining word vectors with machine learning. The approach that implemented GPT-4 (a Large Language Model) demonstrated superior performance, suggesting that Large Language Models are poised to be the preferred method for high-throughput phenotyping of physician notes.}

\section*{Introduction}
The advent of precision medicine has intensified the need for high-throughput phenotyping of electronic health records (EHR). This task remains challenging due to the complexity and volume of physician notes. High-throughput phenotyping—the automated mapping of patient symptoms to standardized ontology concepts—is crucial for this endeavor \cite{sahu2022artificial, afzal2020precision,robinson2012deep,hier2022focused, hier2022dc}. Although traditional and some advanced Natural Language Processing (NLP) methods have progressed toward this goal, their limitations underscore the need for more efficient methods. The emergence of Large Language Models (LLMs) like GPT-4 introduces a promising new approach to address this unsolved problem. This study compares the performance of an LLM, NLP, and hybrid approaches to the high-throughput phenotyping of physician notes from an EHR.

The precision medicine initiative, which aims to match treatments and outcomes with the individual characteristics of each patient, requires computable descriptions of patient signs and symptoms. These descriptions must be both detailed and automated. Despite this critical need for high-throughput methods, their implementation in human medicine has lagged behind other fields, such as agriculture \cite{mir2019high, gehan2017high}. There is a pressing need for more efficient high-throughput methods \cite{alzoubi2019review, pathak2013electronic, shivade2014review}.

Historically, Natural Language Processing (NLP) methods to extract medical concepts from medical text evolved from rule-based and dictionary-based systems \cite{krauthammer2004term, eltyeb2014chemical, quimbaya2016named}. Second-generation systems used machine learning and statistical models to find medical concepts in text \cite{hirschman2002rutabaga, uzuner20112010}. The third generation of approaches to concept extraction saw the application of deep learning methods to this problem, notably RNN (recurrent neural networks) and CNN (convolutional neural networks) \cite{lample2016neural,chiu2016named, habibi2017deep, gehrmann2018comparing, arbabi2019identifying}. The fourth generation systems introduced transformer architecture and BERT (bidirectional encoder representations from transformers), achieving gains in performance due to improvements in attention and language understanding \cite{devlin2018bert, vaswani2017attention, zhu2021utilizing, yu2019biobert, lee2020biobert, ji2020bert}. The emergence of fifth-generation large language models (LLMs)--such as GPT-4 (Generative Pre-Trained Transformer)--offers new flexibility, scalability, and generalizability that have allowed an attack on previously unsolvable NLP problems, including the high-throughput phenotyping of physician notes \cite{yan2023large, yang2023enhancing,wang2023fine}.

Building on a prior pilot study \cite{munzir2024high}, which demonstrated the potential of GPT-4 for high-throughput phenotyping, this study compares three computational approaches on a larger corpus of physician notes:
\begin{enumerate}
\item \textbf{LLM Approach}:  \textit{GPT-4}, that combines a Large Language Model with generative AI capabilities;
\item \textbf{Hybrid Approach}:  \textit{NimbleMiner}, that blends machine learning classification with word vector expansion; and
\item \textbf{NLP Approach}:  \textit{spaCy spancat}, which creates an NLP pipeline that depends on deep learning and word tokenization for span categorization.
\end{enumerate}

This comparison evaluates the performance of these approaches for high-throughput phenotyping, providing insights into their suitability for precision medicine. To establish a ground-truth data set, we manually annotated the signs and symptoms (phenotype) of the patients described in 170 physician notes. We evaluated the accuracy, precision, and recall of the three approaches. The results show the potential for advanced computational approaches to make practical high-throughput phenotyping of EHRs and foreshadow a potential shift towards the dominance of approaches based on large language models.

\begin{mdframed}[
  frametitle={Box 1: Instructions to GPT-4 and the human annotator for finding neurological phenotypes in physician notes},
  frametitlebackgroundcolor=gray!30, 
  backgroundcolor=gray!20, 
  frametitlerule=true,
  frametitlealignment=\centering,
  innertopmargin=10pt,
  innerbottommargin=10pt,
  innerleftmargin=10pt,
  innerrightmargin=10pt,
  roundcorner=10pt,
  font=\footnotesize 
]
Here are instructions for GPT-4 on how to do high-throughput neurological phenotyping. This exercise aims to find neurological signs and symptoms in physician notes and assign each one to one of the 20 phenotype categories below.

\textbf{behavior} [including anxiety, depression, delusion, psychosis, hallucination, etc.]

\textbf{cognitive} [including memory loss, forgetfulness, confusion, cognitive impairment, inattention, dementia, etc.]

\textbf{EOM} [including double vision, abnormal eye movements, diplopia, sixth nerve palsy, third nerve palsy, skew deviation, etc.]

\textbf{fatigue} [including tiredness, lack of energy, poor energy, etc.]

\textbf{gait} [including abnormal gait, spastic gait, ataxic gait, poor balance, imbalance, falling down, falling, using a cane, using a walker, etc.]

\textbf{hyperreflexia} [including increased reflexes, increased biceps reflex, biceps +++, triceps +++, biceps ++++, triceps +++, etc.]

\textbf{hypertonia }[including increased tone, spasticity, hypertonia, muscle spasms, etc.]

\textbf{hyporeflexia} [including decreased reflexes, areflexia, hyporeflexia, bicep 1+ 1+, absent ankle reflex, ankle reflex 1+ 1+, etc.]

\textbf{sphincter }[including urinary frequency, urinary incontinence, constipation, bowel incontinence, urinary retention, etc.]

\textbf{incoordination} [including ataxia, dysmetria, poor coordination, etc.]

\textbf{ON }[including optic neuritis, apd, afferent pupillary defect, pale disk, disk atrophy, on?

\textbf{pain} [including shooting pain, burning pain, allodynia, arm pain, headache, head pain, leg pain, etc.]

\textbf{paresthesias} [including numbness, tingling, loss of sensation, sensory loss, hypesthesia, etc.].

\textbf{seizure} [including seizures, convulsions, fits, attacks, etc.]

\textbf{sleep} [including hypersomnia, insomnia, restless legs, abnormal sleep, trouble sleeping, etc.]

\textbf{speech} [including lack of speech, slurred speech, dysarthria, aphasia, etc.]

\textbf{tremor} [including tremor, tremulousness, action tremor, resting tremor, rubral tremor, etc.]

\textbf{vision} [including impaired vision, decreased visual acuity, visual loss, etc.]

\textbf{weakness} [including weakness, loss of strength, difficulty lifting arms, difficulty lifting legs, biceps 4 4, triceps 4 4, hip flexors 4 4, etc.]

\textbf{cranial nerve} [including any brainstem dysfunction or cranial nerve dysfunction, such as deafness, hearing loss, tinnitus, vertigo]\\

The output should be a list of tuples, where the first element of the tuple is a phenotype category, and the second element is a \textbf{1} if the phenotype was found and a \textbf{0} if it was not. 
\end{mdframed}

\section*{Methods}
\textit{Data Acquisition:} We analyzed physician notes from electronic health records (EHR) of neurology patients diagnosed with multiple sclerosis (MS, ICD-10 code: G35), visiting the University of Illinois at Chicago Neurology Clinic between 2019 and 2022. The physician notes were extracted from the REDCap (Research Electronic Data Capture) system as a CSV file. To avoid analyzing notes with little substantive content, we selected the first physician progress note for each patient that contained at least 600 words.  We excluded discharge summaries, admission notes, and consultation notes.  Non-ASCII characters and quotation marks were removed to facilitate conversion to the JSONL format. Of 547 unique patient records, 188 were used to train the spaCy spancat model (training dataset), and 170 notes were assigned for model evaluation (test dataset).

\textit{Selection of Phenotype Categories:} The 20 phenotype categories (Box 1) were carefully selected based on their frequency in and clinical relevance to multiple sclerosis (MS) \cite{howlett2023subtypes}. Although many granular terms are available to describe neurology phenotypes in the Human Phenotype Ontology, we chose to \say{roll up} these terms into 20 high-level categories to increase the interpretability and comprehensibility of the findings.

\begin{figure}[htbp]
    \centering
    \begin{subfigure}[t]{0.49\textwidth} 
        \centering
\includegraphics[width=\textwidth, valign=t]{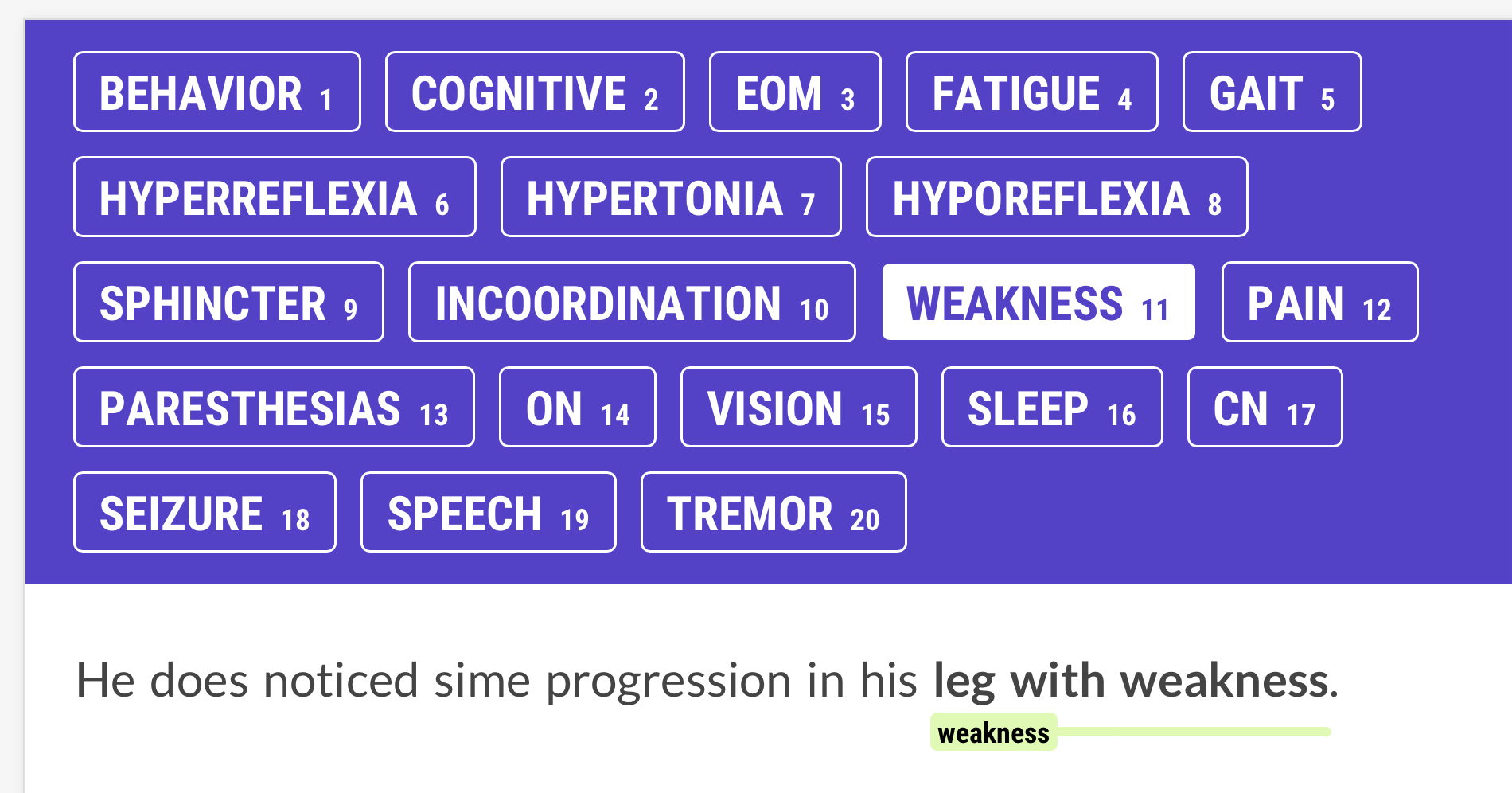}
        \caption{Screen for text span \textit{leg with weakness} annotated as \textbf{weakness}. Note that the physician note contains grammatical and spelling errors.}
        \label{Fig:Weakness as text}
    \end{subfigure}
    \hfill 
    \begin{subfigure}[t]{0.49\textwidth} 
        \centering
        \includegraphics[width=\textwidth, valign=t]{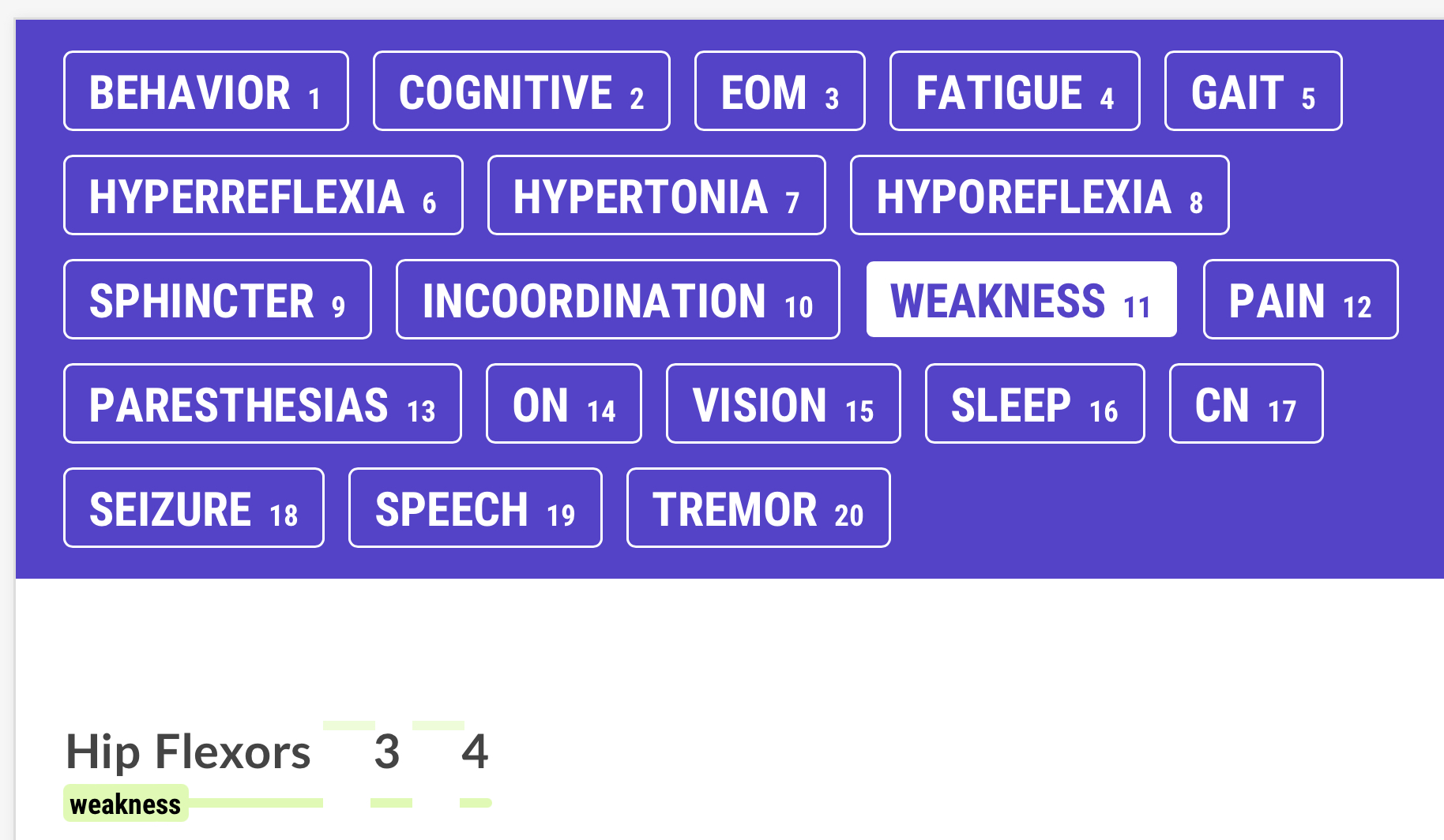}
        \caption{Screen for \textit{Hip Flexors 3 4} annotated as \textbf{weakness}. Note that the physician has coded weakness as a numerical score on a 0 to 5 scale.}
        \label{Fig:Weakness as numeric}
    \end{subfigure}
    \caption{Annotations screens for Prodigy for text spans indicating weakness. The annotator has a choice of 20 labels for selected text spans.}
    \label{Fig:Prodigy_screen}
\end{figure}

\textit{Phenotyping by Human Annotator:} ground-truth labels for the notes were generated using the Prodigy annotation tool (Explosion AI, Berlin). The task involved identifying text spans corresponding to one of 20 neurological signs and symptoms categories. The initial phase of annotations used the \texttt{spans.manual} recipe in Prodigy. After the first ten notes were annotated, a preliminary spaCy spancat model was trained, allowing a switch to the \texttt{spans.correct} recipe, which suggests potential spans (Figure \ref{Fig:Prodigy_screen}). The human annotator received the same instructions as the GPT-4 API (Box 1). Our methods for phenotype annotation have been previously described, including high levels of inter-rater agreement ($\kappa$ = $0.85$) \cite{oommen2023inter}.

\begin{figure}[ht]
    \centering
    \begin{subfigure}[t]{0.4\textwidth} 
        \centering
        \includegraphics[width=\textwidth, valign=t]{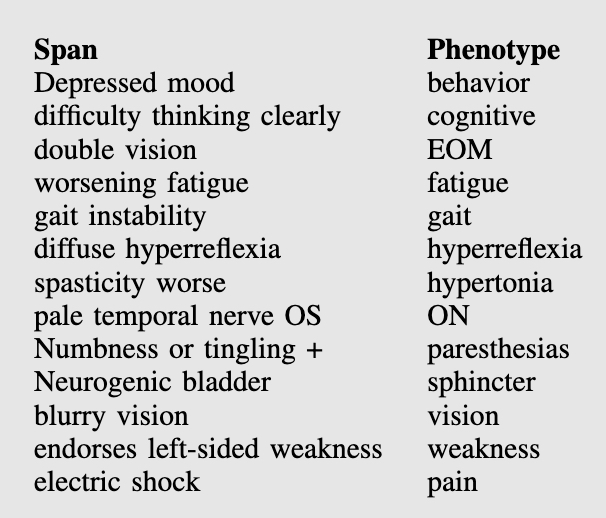} 
        \caption{Sample \textbf{seed terms} passed to NimbleMiner used to generate the simclins utilized by the SVM classifier.}
        \label{fig:NM_sample_phrases}
    \end{subfigure}
    \hfill 
    \begin{subfigure}[t]{0.55\textwidth} 
        \centering
        \includegraphics[width=\textwidth, valign=t]{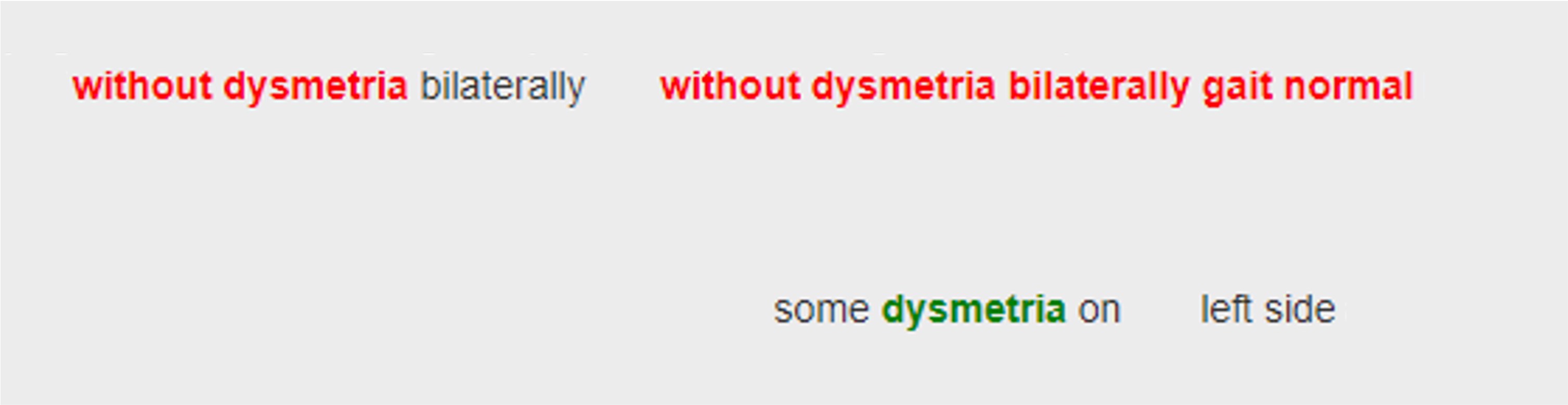} 
        \caption{Example output screen from the Hybrid method (Nimble Miner). NimbleMiner has identified text spans related to the phenotype \say{incoordination}.  \say{Dysmetria} is a simclin for incoordination. NimbleMiner has highlighted two negated examples of dysmetria (red font) and one positive example (green font.)}
        \label{fig:NimbleMiner Dysmetria}
    \end{subfigure}
    \caption{Examples of seed terms used to generate simclins (a) and examples of positive and negated text spans for phenotype identified by NimbleMiner (b).} 
\end{figure}

\textit{Phenotyping by Hybrid Approach:} NimbleMiner \cite{topaz2019nimbleminer} is a tool for the recognition of medical concepts in clinical texts that are implemented in R (compatible with version 4.0.2 \cite{R4.0.2}). It is a hybrid model that combines machine learning classifiers with word embeddings (word2vec) to identify medical concepts. By transforming seed terms into an internal lexicon called \textit{ simclins}, NimbleMiner uses machine learning classifiers to find matching phrases in clinical narratives. NimbleMiner adheres to a label classification strategy that uses only positive labels and excludes negated concepts. An initial list of signs and symptoms of multiple sclerosis was augmented by text spans from neurology notes (Figure \ref{fig:NM_sample_phrases}). NimbleMiner uses prenegations, terms that precede and negate a span (e.g., \textit{ no sign of weakness}), and postnegations, terms that follow and negate a span (e.g., \textit{ weakness negative}) to exclude negated phenotypes. We selected an SVM classifier within NimbleMiner due to the known proficiency of SVM classifiers on the text categorization tasks \cite{joachims1998text}. The SVM classifier determined the binary presence of the 20 neurological phenotypes in the physician notes (Figure \ref{fig:NimbleMiner Dysmetria}). 

\begin{figure}
    \centering
    \includegraphics[width= 0.95\textwidth]{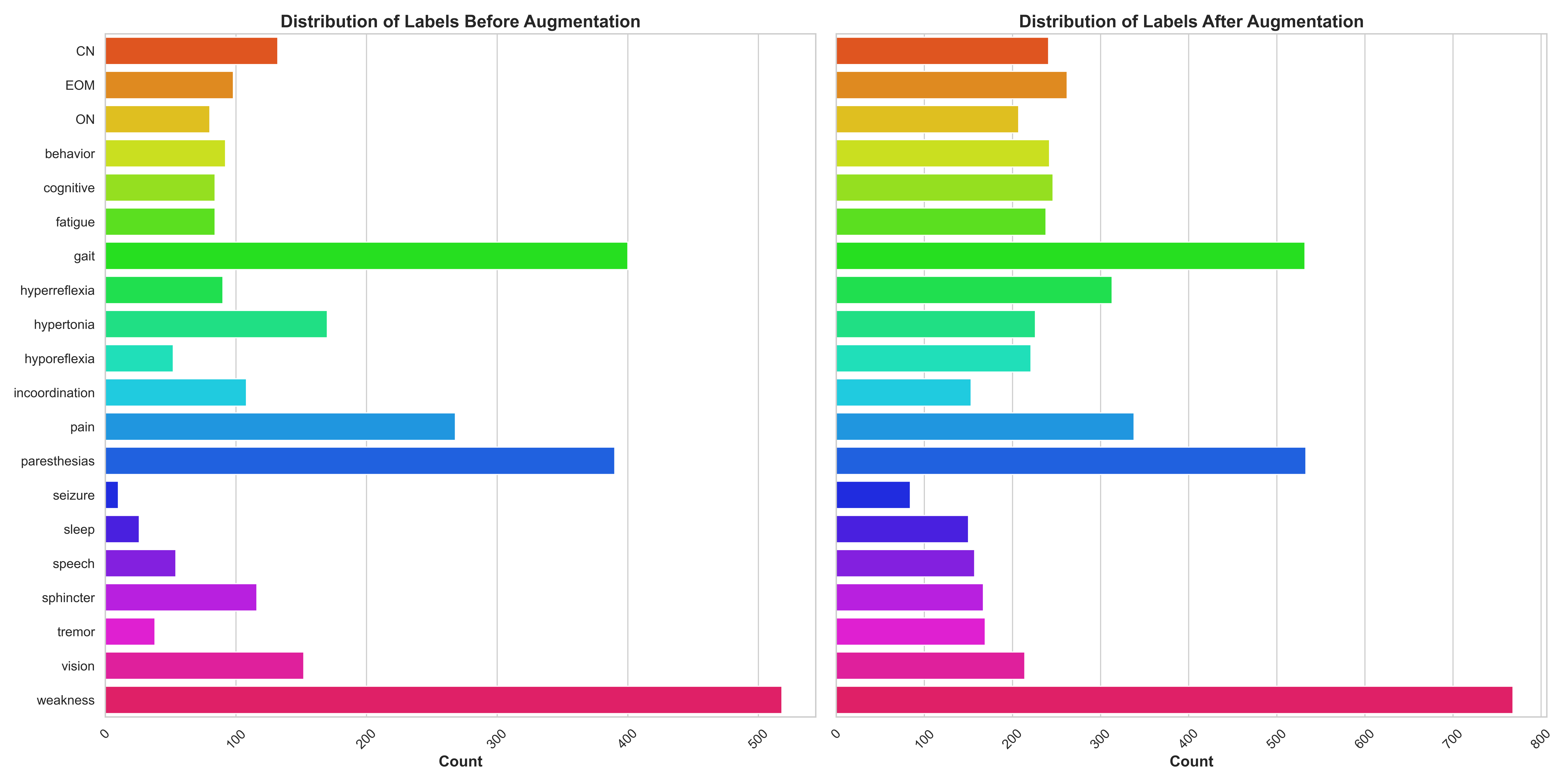}
    \caption{Due to class imbalance in the training dataset for the NLP spancat model, synthetic data was added, increasing the number of lines annotated from 11,688 to 15,052 and thus increasing the minority classes ON (optic neuritis), seizure, sleep, and tremor. In addition, additional training examples were added to the hyperreflexia, hyporeflexia, and weakness classes due to low recall in these classes (see discussion)}
    \label{Fig:Class_Imbalance}
\end{figure}

\textit{Phenotyping by NLP Approach:}  We used the SpaCy NLP spancat pipeline (Explosion AI, Berlin) to recognize the 20 target phenotype labels. We implemented the default \texttt{tok2vec} component for token-to-vector encoding and the default \texttt{spancat} component for span categorization. Initially, we set the parameters in the \texttt{config.cfg }file to their default values, including the system, components, training, batch, and initialization sections. Our initial training dataset for the spaCy spancat pipeline was 188 physician notes with 11,688 annotated lines. We used the \textit{data-to-spacy} recipe from Prodigy (Explosion AI) to create training and validation sets. The initial F score was unsatisfactory at 0.34, and a class imbalance with poor recall in minority classes was recognized (Figure \ref{Fig:Class_Imbalance}). Manually created synthetic data with more examples from the underrepresented speech, seizure, and tremor classes were added. Furthermore, due to the problematic dual representation of hyporeflexia, hyperreflexia, and weakness as text and numeric values (see Figure \ref{Fig:Prodigy_screen}, additional examples of numerically encoded phenotypes were added, resulting in a dataset of 15,052 annotated lines. With data augmentation, the F in the validation dataset increased to 0.62. We then added transfer learning from a previously trained spancat model to achieve an F score of 0.77. Although we attempted to improve spancat performance further by adding external word vectors and a transformer architecture, software version incompatibility prevented us from implementing those improvements. The outputted \texttt{model-best} was applied to the unseen test dataset to find neurological phenotypes in each note.

\begin{figure}[ht]
    \centering
    \begin{subfigure}[t]{0.3\textwidth} 
        \centering
        \includegraphics[width=\textwidth, valign=t]{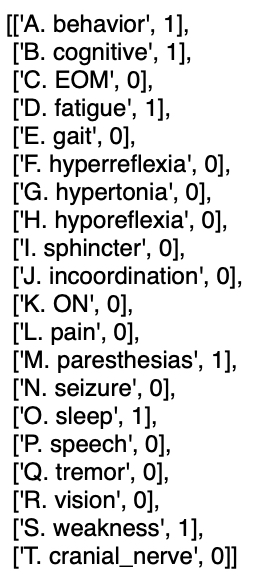} 
        \caption{Example list of phenotype list returned by GPT-4.}
        \label{Fig:gpt4-output}
    \end{subfigure}
    \hfill 
    \begin{subfigure}[t]{0.5\textwidth} 
        \centering
        \includegraphics[width=\textwidth, valign=t]{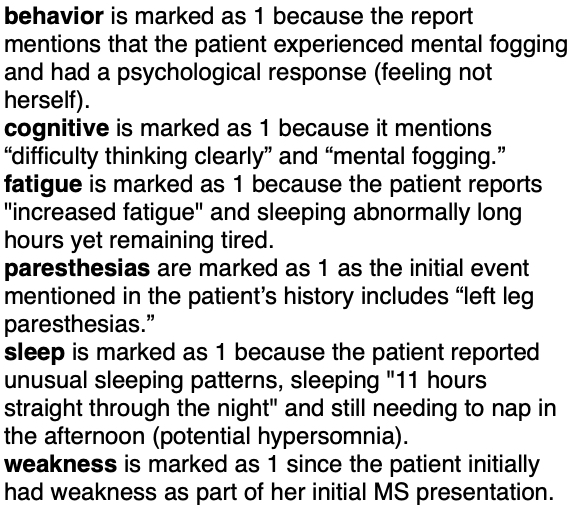} 
        \caption{Explanations provided by GPT-4 for phenotyping.}
        \label{Fig:gpt4-explain}
    \end{subfigure}
    \caption{For physician note, GPT-4 outputted a list of phenotypes (a) and explanations for its choices (b).}
    \label{Fig:gpt4-comparison}
\end{figure}

\textit{Phenotyping by LLM Approach}. The instructions for phenotyping the physician notes were passed to GPT-4 API \cite{OpenAI2024ChatGPT} as a prompt (Box 1). Initial modifications to the prompt were made interactively with GPT-4 in the chat mode. We used the chat mode to resolve ambiguities in the prompt, such as whether to categorize \say {facial weakness} as a finding of \textit{weakness} or a finding of \textit{cranial nerve}. Further prompt modifications were needed to obtain a GPT-4 output that could be parsed into a pandas DataFrame. We wrote a Python script that iterated through the physician notes using the GPT-4 API. No speed or complexity limitations were experienced with the high-throughput phenotyping of the 170 physician notes (each approximately 1,000 words). GTP-4 generated a list of phenotypes (parsable by a Python script) (Figure \ref{Fig:gpt4-output})  and a brief explanation of its choices (Figure \ref{Fig:gpt4-explain}).  The GPT-4 output was saved in a pandas DataFrame for further analysis.

\textit{Calculation of Performance Metrics}.  The ground-truth labels for each physician note were stored in the Prodigy SQLite database. We used Python to convert the ground-truth annotations into a 170 x 21 pandas DataFrame where the first column was the Record ID, and the next 20 columns held the binarized value for each of the 20 phenotypes (present or absent). We created similar 170 x 21 dataframes from the binarized predictions of the NLP, Hybrid, and LLM approaches. To evaluate the performance of the computational approaches for high-throughput phenotyping, we selected precision, recall, and accuracy as our primary metrics. These metrics were chosen for their direct interpretability and specific relevance to the binary classification tasks. Python was used to calculate accuracy, precision, and recall by standard methods \cite{velupillai2009developing, PrecisionAndRecallWiki}. A micro average was calculated for each of the twenty phenotype categories, and an unweighted overall macro average across all categories.

\textit{Human Studies:} The research was approved by the Institutional Review Board of the University of Illinois at Chicago.  All physician notes were unidentified, and all protected health information was deleted. GPT-4 did not retain or use patient health information for LLM training.

\begin{figure}
    \centering
    \includegraphics[width =0.9\textwidth]{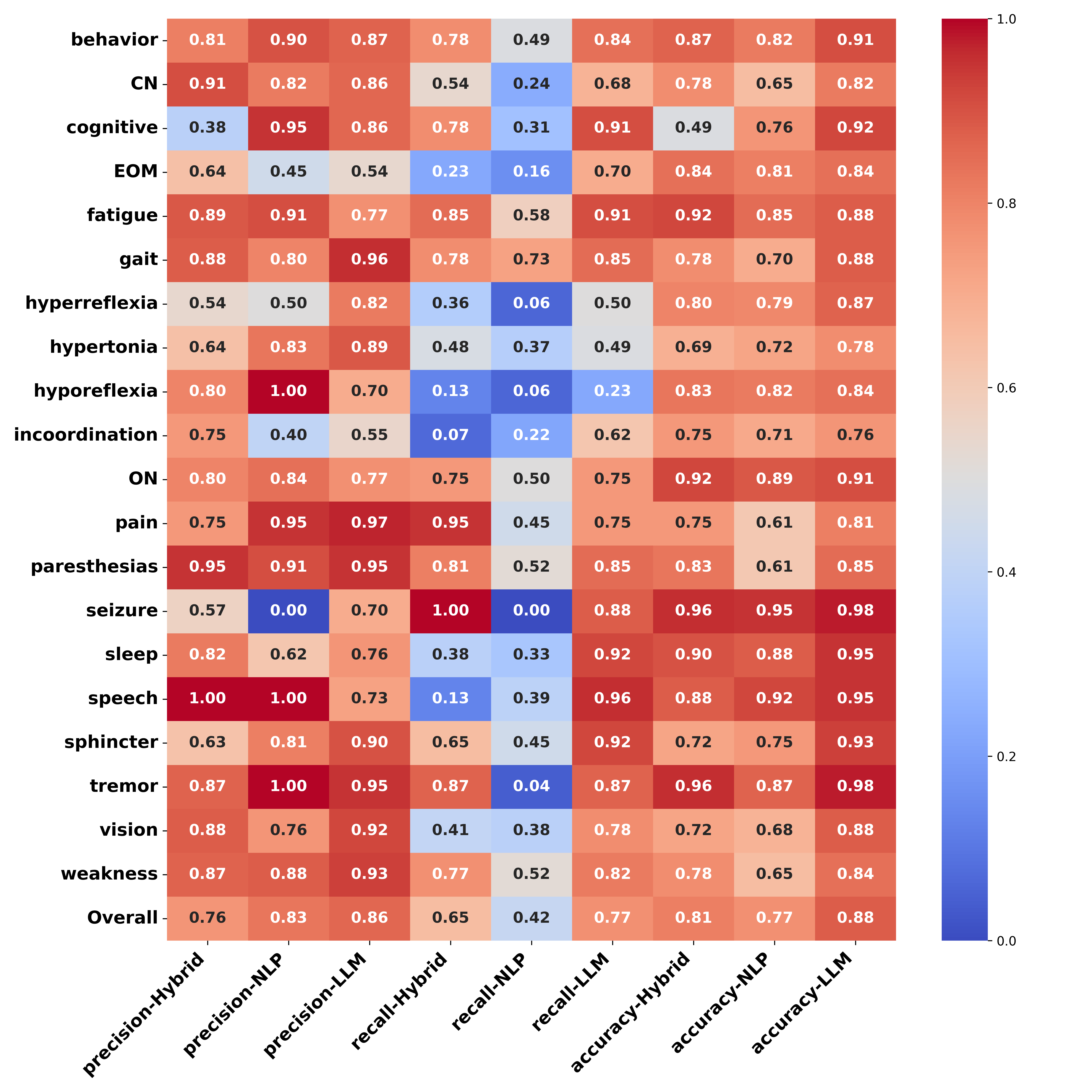}
    \caption{Heat map showing precision, recall, and accuracy for three high-throughput phenotyping approaches: Hybrid, NLP, and LLM. Individual phenotype category metrics are micro averages; the overall metrics are macro averages.  Abbreviations include \textbf{CN} (cranial nerve and brainstem), \textbf{EOM} (extraocular eye movements), and \textbf{ON} (optic neuritis). The category paresthesias includes sensory loss, numbness, and tingling)}.
    \label{Fig:Heat_map_comparison}
\end{figure}

\section*{Results}
\label{s:results}
We compared the performance of three high-throughput phenotyping approaches. After iterative improvements to the Hybrid (NimbleMiner) and the NLP (spaCy spancat) approaches, all showed good accuracy (Figure \ref{Fig:Heat_map_comparison}). The LLM (GPT-4) Approach performed the best (0.88), followed by the Hybrid Approach (NimbleMiner) (0.81) and the NLP Approach (spaCy spancat) (0.78). Precision (with higher scores reflecting lower false positive rates) was high for all three approaches, although the LLM Approach performed best. Recall (with higher scores reflecting lower false negative rates) was higher for the LLM Approach (0.77), lower for the Hybrid Approach (0.65), and lowest for the NLP Approach (0.42).

Both the training dataset (Figure \ref{Fig:Class_Imbalance}) and the test dataset (not shown) demonstrated class imbalances. In particular, the classes of \textit{tremor}, \textit{speech}, and \textit{seizures} were underrepresented. Three phenotype classes (\textit{hyporeflexia}, \textit{hyperreflexia}, \textit{weakness}) were dual encoded by physicians in notes who used either descriptive text or numeric scores, such as \textit{weakness} documented as the text ("leg with weakness") or as the numerical score ("Hip Flexors 3 4") (see Figure \ref{Fig:Prodigy_screen}).

Regarding recall—with a few exceptions—the LLM Approach outperformed the Hybrid and the NLP approaches in almost all phenotype categories, with \textit{pain} and \textit{seizure} being the exceptions. Although the superiority of the LLM Approach in precision for individual phenotypes was less pronounced, it surpassed the Hybrid and NLP approaches in several categories, including \textit{speech}, \textit{sleep}, \textit{ON}, \textit{incoordination}, \textit{fatigue}, and \textit{CN}. The LLM Approach consistently outperformed the other approaches across the macro overall performance metrics, including accuracy, precision, and recall (Figure \ref{Fig:Heat_map_comparison}).

\section*{Discussion}
\label{s:discussion}
High-throughput phenotyping of patient data, crucial to advancing precision medicine, involves converting signs and symptoms from clinical notes into computable codes. Given the number of electronic health records and various linguistic challenges that include synonymy, polysemy, irregular abbreviations, colloquialisms, misspellings, and non-standard terminologies, automated methods are essential but face significant obstacles. To be useful, the phenotyping of text held in EHRs must be fast, accurate, and detailed \cite{robinson2012deep}.

We performed high-throughput phenotyping on 170 physician notes using three computational approaches: Hybrid, NLP, and LLM.  All notes were written by neurologists and carried a diagnosis of multiple sclerosis. Phenotyping involved finding the number of occurrences of 20 categories of neurological symptoms. Since writing styles and habits differ between physicians (some repeat signs and symptoms multiple times in their notes, others do not), the results were binarized so that the occurrence of a neurological sign or symptom (phenotype) was recorded as \say{present} or \say{absent} in each note. 
 
 Hybrid, NLP, and LLM approaches performed at high levels of accuracy (0.81, 0.77, and 0.88, respectively (Figure \ref{Fig:Heat_map_comparison}). These accuracies are impressive given that the level of agreement between human annotators reaches a ceiling at $\kappa$ $\approx$ $0.85$ \cite{oommen2023inter}. The superior performance of the LLM approach is notable given the complexity of this multiclass classification task with high number of classes and class imbalances \cite{grandini2020metrics, aly2005survey}. In particular, the NLP approach (spaCY spancat) faced challenges due to low counts in minority classes (\textit{seizures}, \textit{sleep}, and \textit{EOM}), as shown in Figure \ref{Fig:Heat_map_comparison}. This difficulty was partially addressed by class rebalancing using synthetic data. Dual encoding certain phenotypes as a numerical score and textual description (see Figure \ref{Fig:Prodigy_screen} proved challenging for all approaches but less so for the LLM approach. 

The superior performance of the LLM method on minority phenotype classes (\textit{speech}, \textit{tremor},\textit{ seizure}) and dually encoded phenotype classes (\textit{hyporeflexia}, \textit{hyperreflexia}, \textit{weakness}) is notable. With its superior performance in underrepresented classes and its better performance on dually represented phenotypes, GPT-4 demonstrated the ability to handle class imbalances and decode mixed-format data, likely related to its extensive pre. These results highlight the potential role of the LLM approach in high-throughput phenotyping of physician notes. The extensive pretraining of GPT -4 gave it advantages over the other approaches when analyzing misspelled, irregular, or ambiguous text. Furthermore, the LLM (GPT-4) method offered explanations for its selections without prompting (Figure \ref{Fig:gpt4-comparison}), suggesting that advances in explanatory AI (XAI) \cite{minh2022explainable} had been incorporated into the model architecture. 

Several differences in the ease of implementation between the three approaches should be mentioned. The implementation of the LLM method (GPT-4) was straightforward. We used the GPT-4 chat mode to refine the prompt for high-throughput phenotyping (Box 1).  When we implemented the GPT-4 API, additional changes were needed in the prompt to resolve ambiguities and obtain the appropriate output for conversion to a pandas DataFrame (Figure \ref{Fig:gpt4-output}).  The configuration of the Hybrid approach (NimbleMiner)  required a meticulous selection of \textit{seed terms} for each of the 20 phenotype categories as well as a rigorous curation of the generated \textit{simclins}. We went through several iterations of seed generation and simclin curation until acceptable levels of accuracy were obtained. Implementing the NLP approach (spaCy spancat) was the most time-consuming. We created a training dataset by annotating physician notes for the initial spancat pipeline.  Due to poor model accuracy in minority classes (especially low recall) and class imbalances, additional model training was performed with synthetic data. Further improvements in the performance of the spaCy spancat pipeline depended on the implementation of transfer learning from a previously trained model.  

Implementation of high-throughput neurological phenotyping was easiest with the LLM Approach. Furthermore, the LLM Approach outperformed the NLP and Hybrid approaches in accuracy and recall (Figure \ref{Fig:Heat_map_comparison}). Although our results with the LLM Approach (GPT-4) are encouraging, confirmation of these results with a larger and more diverse corpus of physician notes is needed. Several limitations of this study should be mentioned.
\begin{enumerate}
    \item \textit{High through phenotyping was done with a limited number of neurological notes, all with a diagnosis of multiple sclerosis}.  High-throughput phenotyping on more notes with different diagnoses should be studied.
    \item \textit{The phenotyping was done at a coarse level of detail.} We used 20 broad categories for the phenotyping. However, phenotyping can be performed at a more granular level.  For example, the Human Phenotype Ontology has approximately 7,500 terms to document human phenotypes \cite{kohler2021human}. The ability of the LLM method to phenotype at higher levels of granularity should be studied.
    \item \textit{Additional fine-tuning of the Hybrid Approach (NimbleMiner) would likely have improved accuracy}. Additional seed terms and \textit{simclin} curation could have improved performance in some low-performing categories such as \say{cognitive}, \say{sphincter}, and \say{EOM}.
    \item \textit{Modifications to the NLP Approach (spaCy spancat) would probably have improved performance.} Changes that would likely have improved performance include adding a transformer architecture to the pipeline, adding specialized pre-trained word vectors to the pipeline, additional training examples, and better balancing of the phenotype classes in the training dataset.
\end{enumerate}

This study is an indication of the power, simplicity,  and generalizability of large language model approaches when applied to the high throughput phenotyping of EHRs. Large language models (LLMs) are poised to become the dominant approach to high-throughput phenotyping. GPT-4 outperformed more traditional approaches and proved easier to implement. A broader integration of large language models into electronic health records for phenotyping will depend on additional research that validates these findings with different note types, different EHR data types, and in different medical fields. If large language models are to be used in patient care, a determination of their regulatory status will be needed as well as an evaluation of safety, privacy, and security concerns. An assessment of the accuracy of large language models for high throughput phenotyping using recognized ground-truth datasets is needed. Large language models are a significant advance in high-throughput EHR phenotyping.  Greater accuracy can be expected with additional training and fine-tuning of the underlying models.

\nocite{}
\bibliographystyle{vancouver} 
\bibliography{references} 
\end{document}